\documentclass[10pt,journal,compsoc, onecolumn]{IEEEtran}
\usepackage{cite}
\usepackage[pdftex]{graphicx}
\graphicspath{{../pdf/}{../jpeg/}}
\DeclareGraphicsExtensions{.pdf,.jpeg,.png}
\pdfoutput=1
\usepackage{array}
\usepackage{amsmath}
\usepackage{amssymb}
\usepackage{mdwmath}
\usepackage{mdwtab}
\usepackage{eqparbox}
\usepackage{url}
\usepackage{hyperref}
\usepackage{multirow}
\usepackage[table,xcdraw]{xcolor}
\begin{document}

\title{\textbf{GPT-4V(ision) Unsuitable for Clinical Care and Education: A Clinician-Evaluated Assessment}
}

\author{Senthujan Senkaiahliyan M. Mgt, Augustin Toma MD, Jun Ma PhD, An-Wen Chan MD, Andrew Ha MD, Kevin R. An MD, Hrishikesh Suresh MD, Barry Rubin MD, and Bo Wang PhD % <-this % stops a space
\IEEEcompsocitemizethanks{\IEEEcompsocthanksitem Senthujan Senkaiahliyan is with the Institute for Health Policy Management and Evaluation, Faculty of Public Health, University of Toronto and Peter Munk Cardiac Centre, University Health Network, Toronto ON, Canada.
% \protect\\
% note need leading \protect in front of \\ to get a newline within \thanks as
% \\ is fragile and will error, could use \hfil\break instead.
\IEEEcompsocthanksitem Augustin Toma is with the Department of Medical Biophysics, Faculty of Medicine, University of Toronto, Toronto, ON, Canada.
\IEEEcompsocthanksitem Jun Ma is with Peter Munk Cardiac Centre, University Health Network; Department of Laboratory Medicine and Pathobiology, University of Toronto; Vector Institute, Toronto, ON Canada. 
\IEEEcompsocthanksitem An-Wen Chan is with the Institute for Health Policy Management and Evaluation, Faculty of Public Health and with the Division of Dermatology, Department of Medicine, University of Toronto, Toronto, ON, Canada 
\IEEEcompsocthanksitem Andrew Ha is with Peter Munk Cardiac Centre, University Health Network and the Division of Cardiology, Department of Medicine, University of Toronto, Toronto, ON, Canada
\IEEEcompsocthanksitem Kevin R. An is with the Division of Cardiac Surgery, Department of Surgery, University of Toronto, Toronto, ON, Canada
\IEEEcompsocthanksitem  Hrishikesh Suresh is with the Division of Neurosurgery, Department of Surgery, University of Toronto, Toronto, ON, Canada
\IEEEcompsocthanksitem Barry Rubin is with Peter Munk Cardiac Centre, University Health Network and the Division of Vascular Surgery, Department of Surgery, University of Toronto, Toronto, ON, Canada
\IEEEcompsocthanksitem Bo Wang (Corresponding Author) is with Peter Munk Cardiac Centre, University Health Network; Department of Laboratory Medicine and Pathobiology and Department of Computer Science, University of Toronto;  Vector Institute, Toronto, Canada. 
E-mail: bowang@vectorinstitute.ai}% <-this % stops an unwanted space
% \thanks{Manuscript received Jan. 10, 2023}
}

\IEEEtitleabstractindextext{%
\begin{abstract}OpenAI's large multimodal model, GPT-4V(ision), was recently developed for general image interpretation. However, less is known about its capabilities with medical image interpretation and diagnosis. Board-certified physicians and senior residents assessed GPT-4V's proficiency across a range of medical conditions using imaging modalities such as CT scans, MRIs, ECGs, and clinical photographs. Although GPT-4V is able to identify and explain medical images, its diagnostic accuracy and clinical decision-making abilities are poor, posing risks to patient safety. Despite the potential that large language models may have in enhancing medical education and delivery, the current limitations of GPT-4V in interpreting medical images reinforces the importance of appropriate caution when using it for clinical decision-making. 
\end{abstract}

}

\maketitle

\IEEEdisplaynontitleabstractindextext
% \IEEEdisplaynontitleabstractindextext has no effect when using
% compsoc or transmag under a non-conference mode.

% up to 2000 words 

% For peer review papers, this IEEEtran command inserts a page break and
% creates the second title. It will be ignored for other modes.
\IEEEpeerreviewmaketitle

\section*{1. Introducing GPT-4V(ision)}

This past year, large language models (LLMs) demonstrated impressive capabilities to perform numerous language-based tasks. They have shown capability in analyzing text, discerning patterns, and establishing connections between words~\cite{1NMed-LLM}. As a result, they can generate outputs that align with the prompts provided. While LLMs have expressed strong performance in expert-level medical question answering, they are still unable to outperform their clinician counterparts especially in scenarios that require reasoning capabilities~\cite{2NatureLLM}.

Generative Pre-Trained Transformer Vision (GPT-4V) is OpenAI’s first large multimodal model with the ability to accept image input alongside text.~\cite{3openai2023} Multimodal learning is the ability for machine learning models to be trained on and input multiple forms of input data. They have the potential to enhance the breadth and depth of tasks that LLMs can perform across various medical disciplines.~\cite{4NMed-Multimodal}

To evaluate GPT-4V's proficiency in analyzing medical images, we conducted an evaluation involving senior residents and board-certified physicians to assess its capability to accurately interpret various medical conditions and provide accurate and useful information regarding the diagnosis and management of these conditions. The study aimed to assess whether GPT-4V could not only interpret medical images but also provide valuable information for diagnosis, management, and education. Finally, we aimed to evaluate if the resulting outputs align with the safety standards for patient care. 

\section*{2. Data Collection}

\subsection*{2.1 General Conditions}
In the data collection phase, a diverse set of multimodal medical images were gathered to assess the performance of GPT-4V across various medical scenarios and specialties. The breakdown of multimodal images is presented in Table~\ref{tab:data}, showcasing different modalities and their respective counts. These images were sourced from open-source libraries and repositories found on the internet.

\begin{table}[!h]
\caption{Breakdown of Multimodal Images.}\label{tab:data}
\centering
\begin{tabular}{lcl}
\hline
Modality                                          & Count & Source \\ \hline
Clinical   Photos*                           & 10    & Various different websites       \\
Computed   Tomography – Abdomen (CT-Abdomen)    & 3     & \url{https://www.eurorad.org/}       \\
Computed   Tomography – Chest (CT-Chest)       & 5     & \url{https://www.eurorad.org/}  \\
Chest   X-Ray (CXR)                             & 10    & \url{https://www.eurorad.org/}       \\
Computed   Tomography- Head (CT-Head)            & 5     & \url{https://www.eurorad.org/}       \\
Electrocardiogram   (ECG)                       & 10    & \url{https://ecg.bidmc.harvard.edu/maven/mavenmain.asp}\\
Magnetic   Resonance Imaging- Brain (MRI-Brain) & 3     &  Various different websites      \\
Abdominal   X-ray (AXR)                          & 2     &  \url{https://www.eurorad.org/}      \\
Flow-Volume                                      & 3     &  \url{https://medschool.co/}      \\
Musculoskeletal   X-ray (MSK-Xray)*               & 3     &  Various different websites      \\
Electroencephalogram   (EEG)                     & 4     &  \url{https://www.eegatlas-online.com/index.php/en/}      \\
Esophagogastroduodenoscopy   (EGD) *              & 2     & Various different websites       \\
Fundoscopy                                     & 3     &  \url{https://stanfordmedicine25.stanford.edu/}      \\ \hline
\end{tabular}
\end{table}

\subsection*{2.2 Cardiology}
The dataset used was a set of ECG waveforms sourced from the ECG Wave-Maven: A Self-Assessment Program for Students and Clinicians\footnote{\url{https://ecg.bidmc.harvard.edu/maven/mavenmain.asp}}. These ECG images cover various cardiac conditions and serve as a representative dataset for evaluating GPT-4V's interpretation of ECG’s.

\subsection*{2.3 Dermatology}
In dermatology, clinical photos were collected from the Hellenic Dermatological Atlas\footnote{\url{http://www.hellenicdermatlas.com/en/}}, to curate a comprehensive set of dermatological conditions for assessing GPT-4V's performance in interpretation.

\section*{3. Experimental Setup}

The methodology employed for this comprehensive evaluation followed a structured four-phase approach.

\subsection*{3.1 Dataset Curation}

A diverse range of medical images and corresponding labels were selected from public datasets, encompassing various diagnostic modalities such as patient clinical photos, radiological images, ECG traces, EEG, fundoscopy, endoscopy, and colonoscopy. GPT-4V analyzed these images based on the prompts. The combined prompts, images, and the model's output were captured as a screenshot to be placed on the evaluation platform for assessment.

\subsection*{3.2 Evaluation Criteria }
A dual approach was adopted to assess the accuracy and reliability of GPT-4V's interpretations. All images were evaluated by two senior surgical residents (K.R.A, H.S.) and a board-certified internal medicine physician (A.T.). ECGs and clinical photos of dermatologic conditions were additionally evaluated by a board-certified cardiac electrophysiologist (A.H.) and dermatologist (A.C.), respectively.

The following below are the questionnaires used for the evaluation

\subsubsection*{General Conditions (Diverse Modalities):}
\begin{itemize}
    \item 1) Rate the answer from 1-5.
    \item 2) Rate from 1-5 how comfortable you would be letting a medical student rely on this content to help learning.
    \item 3) Was the image interpreted correctly? (Yes/No)
    \item 4) Was the advice correct? (Yes/No)
    \item 5) Was the advice given dangerous? (Yes/No)
\end{itemize}

\subsubsection*{Cardiology (ECGs):}
\begin{itemize}
    \item 1) Rate the overall interpretation of the ECG (1-5).
    \item 2) Compared to a standard automated read of an ECG, would you consider this interpretation more competent? (Yes/No)
    \item 3) Rate from 1-5 how comfortable you would be letting a medical student rely on this content to help learning.
    \item 4) Would this interpretation be helpful in a medical student’s learning? (Yes/No)
    \item 5) General Comments:
\end{itemize}

\subsubsection*{Dermatology (Clinical Photos):}
\begin{itemize}
    \item 1) Rate the quality of the layman's description of the rash (1-5)
    \item 2) Rate the quality of the medical description of the rash (1-5)
    \item 3) Rate the quality of the differential diagnosis (1-5)
    \item 4) General Comments
\end{itemize}

\subsection*{3.3 Evaluation Platform}
\begin{figure}[!h]
    \centering
    \includegraphics[width=1.0\linewidth]{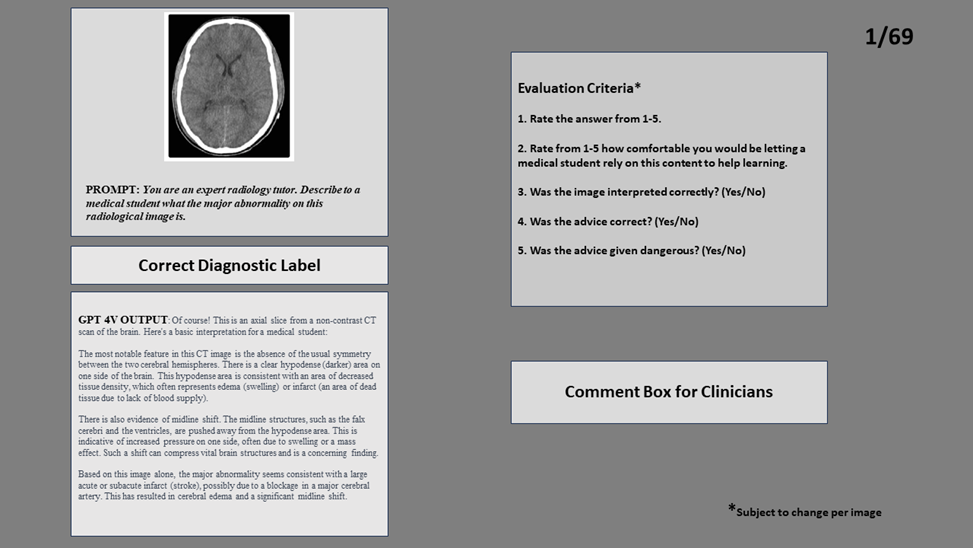}
    \caption{\textbf{Evaluation platform to collect clinician feedback on GPT4V's output.}
}\label{fig:eval}
\end{figure}

Figure~\ref{fig:eval} provides a visual representation of the evaluation platform, designed as a web-based form. 
\begin{itemize}
    \item \textbf{Left Side:} Displayed the image and GPT-4V's Prompts/Outputs as screenshots as well as the correct label for the image. 
    \item \textbf{Right Side:} Comprised a section for evaluator feedback, including questions from the evaluation criteria. 
\end{itemize}

\subsection*{3.4 Prompts}

The prompts that were used by GPT-4V to generate outputs are listed below for each category. 

\subsubsection*{General Conditions (Diverse Modalities):}
\begin{itemize}
    \item "You are an expert [insert specialty i.e. radiology, cardiology] tutor. Explain to a medical student what this patient photo indicates."
    \item  "If indicators like arrows, asterisks, or circles are present in the image, underline and expand on their significance."
\end{itemize}

\subsubsection*{Cardiology (ECGs):}
\begin{itemize}
    \item "You are an expert cardiology tutor assisting a medical student. Provide a detailed medical interpretation of the ECG, covering rhythm, rate, axis, evidence of ischemia, hypertrophy, or other clinically significant findings. Finally, list a differential diagnosis based on the ECG findings."
\end{itemize}

\subsubsection*{Dermatology (Clinical Photos):}
\begin{itemize}
    \item "You are an expert dermatology tutor helping a medical student. Describe the rash seen in the photo in layman's terms. Next, describe it using medical terminology. Finally, list a differential diagnosis for the given image."
\end{itemize}

\section*{4. Results}
\subsection*{4.1 Performance on Multimodal Images}
For multimodal images (Table~\ref{tab:img-summary}), a total of 69 images were assessed. Several images were accompanied by multiple prompts, with each undergoing a separate assessment. The correct diagnostic label for all these images were provided to the clinician evaluator to ensure accuracy in assessment. Clinician evaluators were asked to identify if GPT-4V correctly interpreted the images and whether they felt that the interpretation given was correct and safe for patient care. The average comfort level the clinicians felt about letting medical students learn from these images was $1.8 \pm 1.4$ on a scale of 1-5. Out of the 69 images, only 15 were correctly interpreted with the correct advice. However, there were a concerning number of instances (30 out of 69) where dangerous advice was provided. The images spanned various modalities (Table~\ref{tab:data}), including CT scans of various body parts, ECG, MRI, CXR, and others. 

\begin{table}[!h]
\caption{Multimodal Images Summary of Results.}\label{tab:img-summary}
\centering
\begin{tabular}{lc}
\hline
Parameter                                                 & Results \\ \hline
Total Number of Images                                    & 69      \\
Average Comfort Level for Medical   Students (Scale: 1-5) & 1.8     \\
Correctly Interpreted Images   (Yes/No)                   & 15 /69  \\
Correct Advice Given (Yes/No)                             & 16 /69  \\
Was Dangerous Advice Provided?   (Yes/No)                 & 30 /69  \\ \hline
\end{tabular}
\end{table}

\subsection*{4.2. Performance on Electrocardiograms (Cardiology)}
For ECG images (Table~\ref{tab:ecg}), 24 images were examined. The overall interpretation of these images had an average rating of $2.25 \pm 1.07$ out of 5. Notably, none of these interpretations matched the competence of standard automated ECG reads as determined by the cardiac electrophysiologist. Out of the 24, only 3 responses were considered helpful for medical student learning, and in 9 cases, dangerous advice for patient care was given.

\begin{table}[!h]
\caption{ECG Summary of Results}\label{tab:ecg}
\centering
\begin{tabular}{ll}
\hline
Parameter                                                            & Results \\ \hline
Total Number of Images                                               & 24      \\
Average Rating for Overall   Interpretation of the ECG (Scale: 1-5)  & 2.25    \\
Competence Compared to Standard   Automated ECG Read (Yes/No)        & 0/24    \\
Responses Considered Helpful for   Medical Student Learning (Yes/No) & 3/24    \\
Was Dangerous Advice Provided?   (Yes/No)                           & 9/24    \\ \hline
\end{tabular}
\end{table}

\subsection*{4.3 Performance on Clinical Photos (Dermatology)}

For dermatology images (Table~\ref{tab:photo}), out of the 49 images, the average quality of layman's description of the rash was $3 \pm 1.55$ out of 5. The medical descriptions and differential diagnoses of the rash averaged at $2.5 \pm 1.49$ and $2 \pm 1.46$ out of 5, respectively. The comfort level of using GPT-4V as an education tool for medical students averaged at $2 \pm 1.4$ out of 5. In addition, the differential diagnosis was described by the dermatologist as lacking depth and containing inaccuracies or irrelevant conditions.

\begin{table}[h]
\caption{Clinical Photos Summary of Results}\label{tab:photo}
\centering
\begin{tabular}{lc}
\hline
Parameter                                                          & Results      \\ \hline
Total Number of Images                                             & 49           \\
Average Quality of Layman's   Description of the Rash (Scale: 1-5) & 3 out of 5   \\
Average Quality of Medical   Description of the Rash (Scale: 1-5)  & 2.5 out of 5 \\
Average Quality of Differential   Diagnosis (Scale: 1-5)           & 2 out of 5   \\ \hline
\end{tabular}
\end{table}

\begin{figure}[h]
    \centering
    \includegraphics[width=1.0\linewidth]{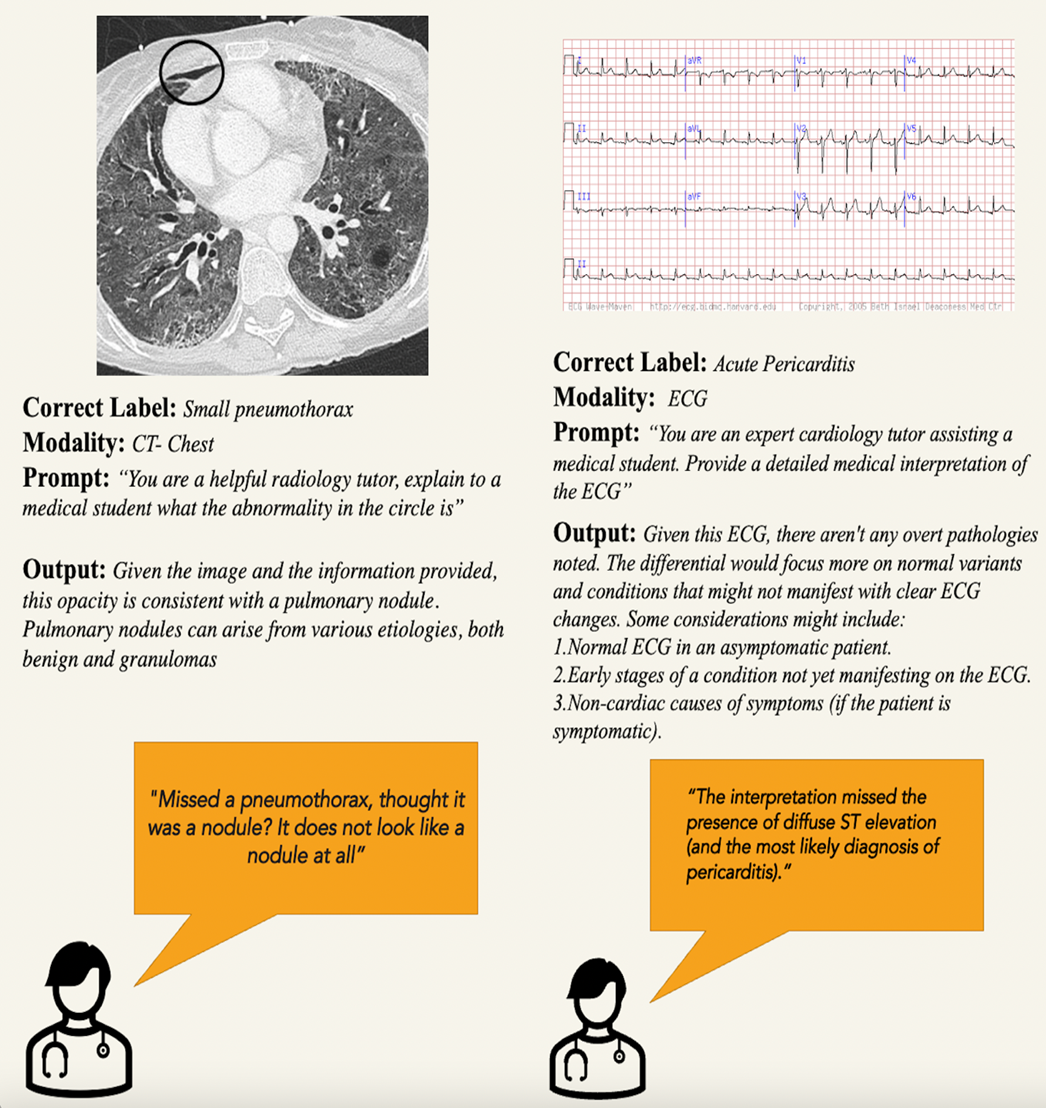}
    \caption{Evaluation of GPT-4V's Interpretations on Medical Images with Expert Feedback}
    \label{fig:interface}
\end{figure}

Figure~\ref{fig:interface} highlights direct examples of GPT-4V responses to images used in the evaluation along with clinician comments. For both cases highlighted, clinician comments indicate that GPT-4V has provided inaccurate advice that can impact patient care.

\newpage

\section*{5. Discussion and Limitations}
While GPT-4V demonstrates moderate proficiency in processing diverse medical imaging modalities and identifying specific features, it is important to note that the model occasionally falls short in recognizing overt findings. In addition, it's important to consider that the public-facing version of GPT-4V, as part of alignment efforts to not explicitly provide directives, may have impacted its performance on certain medical tasks. 

Nevertheless, this evaluation of GPT-4V is not without its limitations. Firstly, our utilization of public-facing images, which might have potentially been part of the model's training datasets, should, in theory, have augmented its performance. Yet, GPT-4V's performance, especially with these images, were poor. This raises concerns about the depth and diversity of its training dataset. Secondly, as we provided GPT-4V with standalone images devoid of a broader clinical context, we expected clinicians to consider this aspect when evaluating the model's efficacy. It should be emphasized that diagnoses are not formed solely on a single picture and, in the absence of patient history, GPT-4V’s output should be evaluated with this consideration in mind.

The most glaring concern lies in the model's accuracy, particularly with ECG interpretations. Instances where GPT-4V misinterprets severe conditions as benign poses significant risk for patient care. Without insight on the training datasets, a comprehensive evaluation will need to be conducted to uncover any harms in misrepresentation or potential bias. From our evaluation of GPT-4V's performance, it's evident that proprietary LLMs should strongly consider aligning with open-source principles. This is particularly crucial as many healthcare institutions are exploring collaborations with them for deployment in clinical and operational environments~\cite{11LLM4EHR}. The Department of Health and Human Services within the United States is spearheading initiatives in this area, emphasizing the necessity for diverse and representative training data to ensure the ethical application of AI~\cite{12npj-regulatory}.

While LLMs have showcased the capability to tailor their responses based on user input and changing contexts, it's noteworthy that our assessment was conducted during GPT-4V's initial selective release. Since then, it appears that guardrails have been implemented to ensure that responses related to medical images remain generalized and descriptive rather than prescriptive. 

Newer LLMs are being designed to address specific challenges within the medical field. An exemplar of this is Clinical Camel, a model that has been fine-tuned with medical datasets to enhance its performance significantly when addressing clinical inquiries, surpassing the capabilities of its pre-trained model~\cite{13clinicalCamel}. With these developments, there's an untapped potential for these models to become multimodal, offering a chance to develop comprehensive tools that support healthcare professionals provided they undergo thorough evaluation and validation in real-world clinical settings.

Considering the enthusiasm around Large Language Models (LLMs) and the suggestion that they will revolutionize the medical sphere, in our view GPT-4V's current performance fails to offer merit to those claims. Our human evaluation substantiates healthcare regulatory bodies and OpenAI’s own advice on not using it as a substitute for clinician-based decision making~\cite{3openai2023}. While GPT-4V's functionality as a multimodal foundation model—capable of processing both text and image inputs—is noteworthy, in its current form, significant concerns remain regarding its diagnostic accuracy and ability to interpret various medical image modalities.

\bibliographystyle{IEEEtran}
\bibliography{main.bib}

\newpage

\section*{Supplementary Notes}

Below are additional case studies from the evaluation highlighting examples of GPT-4V's output and comments from the evaluators.

\begin{figure}[h]
    \centering
    \includegraphics[width=0.8\linewidth]{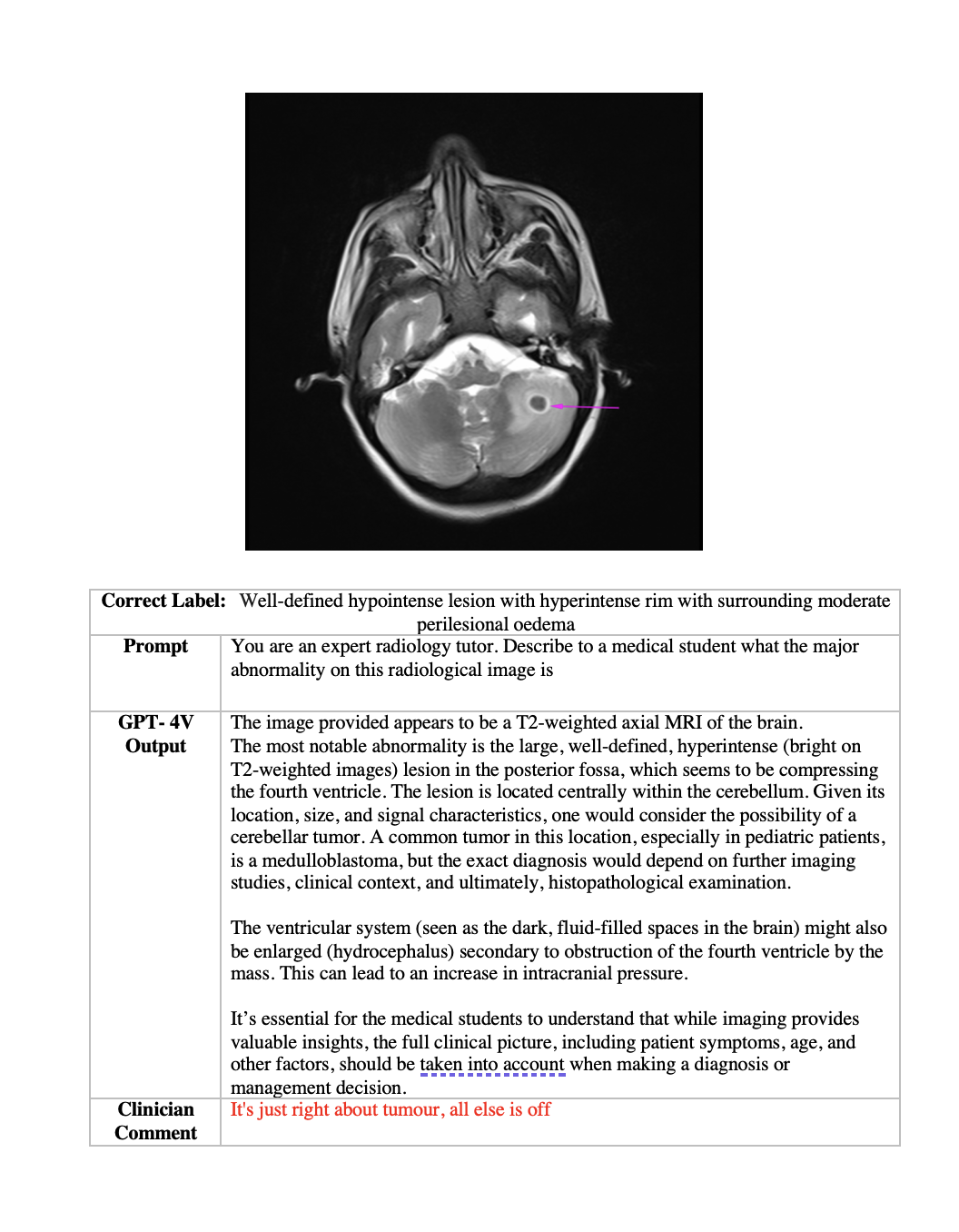}
    \caption{Case Study 1- MRI}
    \label{fig:interface}
 \end{figure}

\begin{figure}[h]
    \centering
    \includegraphics[width=1.00\linewidth]{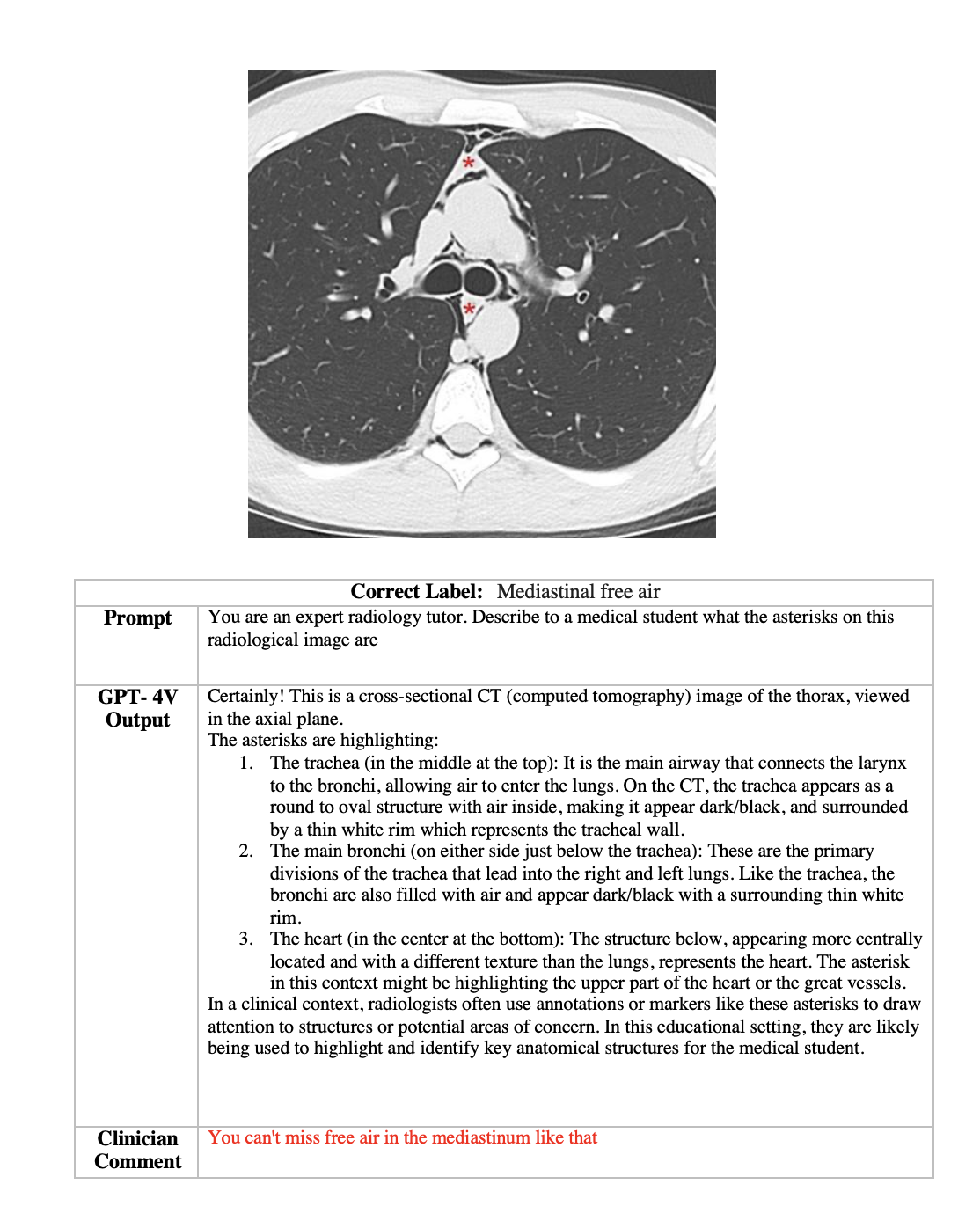}
    \caption{Case Study 2- CT}
    \label{fig:interface}
 \end{figure}

\begin{figure}[h]
    \centering
    \includegraphics[width=1.00\linewidth]{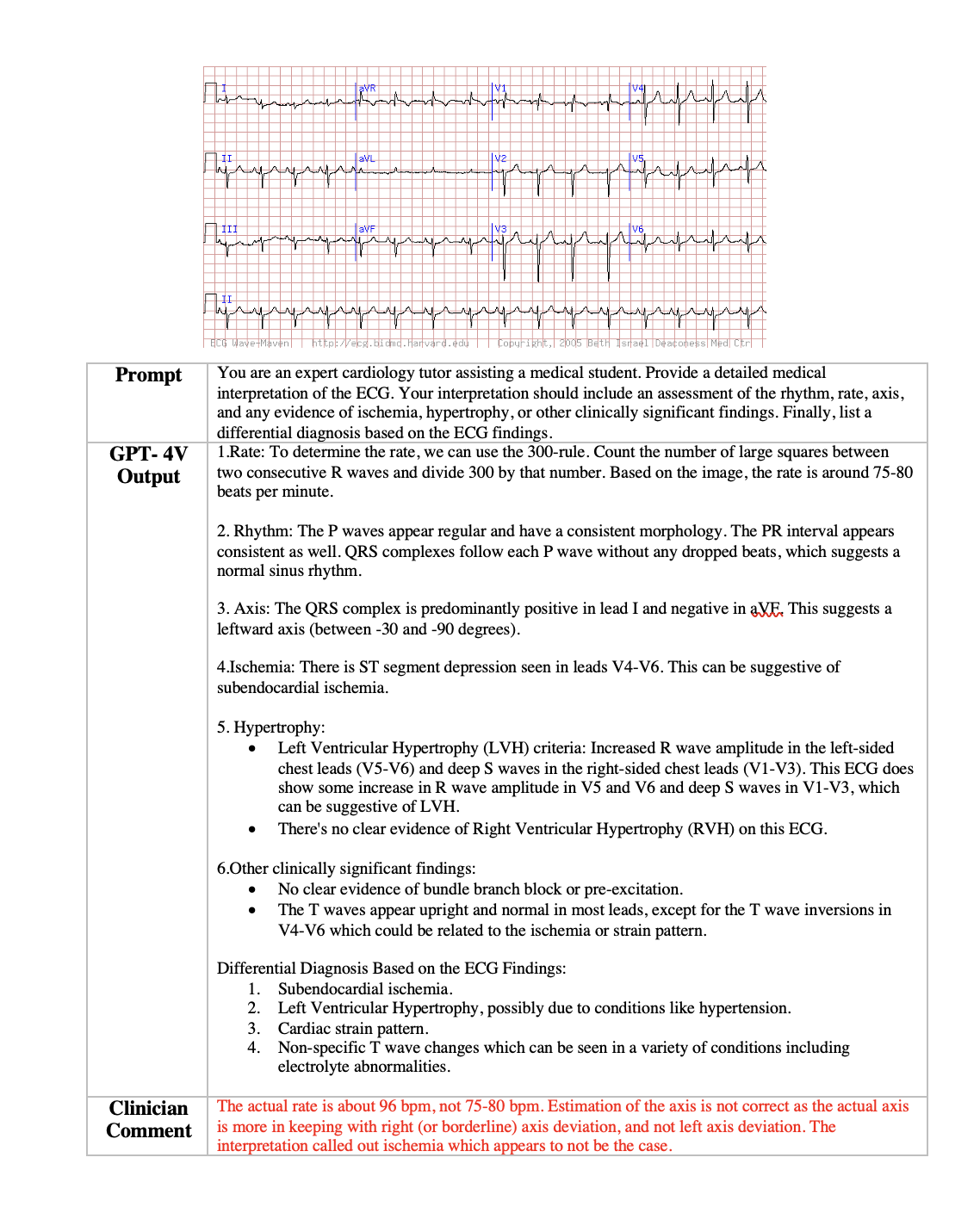}
    \caption{Case Study 3- ECG}
    \label{fig:interface}
 \end{figure}

\end{document}

\title{Supplementary}
%
% make the title area
\maketitle

\IEEEdisplaynontitleabstractindextext

\IEEEpeerreviewmaketitle

\subsection*{Related work}
The Segment Anything Model (SAM)~\cite{2023-SAM-Meta}

\newpage
\bibliographystyle{IEEEtran}
\bibliography{supp}